\documentclass[letterpaper, 10 pt, conference]{ieeeconf}  

\IEEEoverridecommandlockouts                              

\usepackage{algorithm}
\usepackage{algpseudocode}
\usepackage{graphics}
\usepackage{cite}
\usepackage{amsmath,amssymb,amsfonts}
\usepackage{subfig}
\usepackage{textcomp}
\usepackage{xcolor}
\usepackage{nicefrac}       
\usepackage[pdftex]{graphicx}
\usepackage{multirow}
\usepackage{xspace}
\usepackage{lipsum}
\usepackage{booktabs}
\usepackage{tabu}
\usepackage{siunitx}
\usepackage{authblk}
\usepackage{mathtools}

\newcommand{\R}{\ensuremath{\mathbb R}}
\newcommand{\set}[1]{\left\{#1\right\}}
\usepackage{calrsfs}
\DeclareMathAlphabet{\pazocal}{OMS}{zplm}{m}{n}

\title{\LARGE \bf
Event-Based Control for Online Training of Neural Networks
}

\author{Zilong Zhao$^{1}$, Sophie Cerf$^{1}$, Bogdan Robu$^{1}$, and Nicolas Marchand$^{1}$
	\thanks{$^{1}$ Univ. Grenoble Alpes, CNRS, Grenoble INP, GIPSA-lab, 38000 Grenoble, France
		{\tt\small firstname.lastname@gipsa-lab.fr}}%
}

\makeatletter
\def\endthebibliography{%
  \def\@noitemerr{\@latex@warning{Empty `thebibliography' environment}}%
  \endlist
}
\makeatother
\newcolumntype{L}[1]{>{\raggedright\let\newline\\\arraybackslash\hspace{0pt}}m{#1}}
\newcolumntype{C}[1]{>{\centering\let\newline\\\arraybackslash\hspace{0pt}}m{#1}}
\newcolumntype{R}[1]{>{\raggedleft\let\newline\\\arraybackslash\hspace{0pt}}m{#1}}
\begin{document}
\maketitle

\begin{abstract}
Convolutional Neural Network (CNN) has become the most used method for image classification tasks. During its training the learning rate and the gradient are two key factors to tune for influencing the convergence speed of the model. Usual learning rate strategies are time-based i.e. monotonous decay over time. Recent state-of-the-art techniques focus on adaptive gradient algorithms i.e. Adam and its versions. In this paper we consider an online learning scenario and we propose two Event-Based control loops to adjust the learning rate of a classical algorithm E (Exponential)/PD (Proportional Derivative)-Control. The first Event-Based control loop will be implemented to prevent sudden drop of the learning rate when the model is approaching the optimum. The second Event-Based control loop will decide, based on the learning speed, when to switch to the next data batch. Experimental evaluation is provided using two state-of-the-art machine learning image datasets (CIFAR-10 and CIFAR-100). Results show the Event-Based E/PD is better than the original algorithm (higher final accuracy, lower final loss value), and the Double-Event-Based E/PD can accelerate the training process, save up to 67\% training time compared to state-of-the-art algorithms and even result in better performance. 

\end{abstract}

\section{Introduction}
\label{sec:Introduction}
Convolutional Neural Network (CNN) is a popular machine learning algorithm for image classification because it outperforms any other network architecture on visual data. In this paper, we focus on an online learning scenario where data used for training the CNN comes in batches over time \cite{2016arXiv161001030T,hong2015online}. 
A CNN model is a neural network structure with a set of weights which are iteratively learned from training data using methods such as Stochastic Gradient Descent (SGD).
The SGD algorithm is parametrized with a learning rate $\lambda$. A large $\lambda$ helps the model to converge faster but increases the risk of diverging \cite{Bengio2012}. A small $\lambda$ slows the convergence but may lead to a local minimum. 

There are two main learning rate evolution strategies: time-based or adaptive. In most time-based learning rate strategies, $\lambda$ decreases following a predefined decay function \cite{W8305126}. Cyclical strategies have also been developed, where two boundaries are defined and $\lambda$ cyclically varies between them. The disadvantage of these algorithms is that the learning rate path is fixed before training, it cannot be adjusted when necessary. 

Adaptive learning rate algorithms such as Adam \cite{adam2014arXiv1412.6980K}, Nadam (Adam with Nesterov momentum) \cite{dozat2016incorporating} and AMSGrad \cite{amsgradj.2018on} are recent state-of-the-art algorithms which mainly focus on the convergence speed. Different from SGD which uses only the current value of the gradient to update weights, these algorithms use squared gradient to scale the learning rate and take advantage of momentum by using moving average of the gradients. Nevertheless, Wilson et al. \cite{WilsonRSSR17nips} suggested that adaptive gradient methods do not generalize as well as SGD. These methods tend to perform well in the initial portion of training but are outperformed by SGD at later stages of training \cite{DBLP:journals/corr/abs-1712-07628}. To address this issue, AdaBound \cite{adabound} employs dynamic bounds on learning rates to achieve a gradual and smooth transition from adaptive methods to SGD.

Up to our knowledge, E (Exponential)/PD (Proportional Derivative) control \cite{zhao:hal-02115916} is the first adaptive learning rate algorithm which uses control theory to dynamically adapt the learning rate during the learning process. It uses only current gradient as in SGD, but its learning rate $\lambda$ is dynamically calculated based on the loss value. During the E phase, that corresponds to the beginning of the training when the loss value is continuously decreasing, $\lambda$ is increased each time step by a factor of two. Once the loss stops decreasing, the PD phase takes over and, considering CNN as a dynamic system, computes the control input (i.e. $\lambda$) based on the CNN's output (i.e. the loss value). 

The above-mentioned algorithms 
are time-based, in the sense of a periodic computation of the control law regardless its utility. In this paper, we propose two event-based control strategies to reduce the time CNN spends learning "inefficiently" from data, as well as an extensive evaluation. Moreover, while using event-based mechanisms we should expect for a reduction in the use of resources \cite{Astrom:2008,Durand2009c}, without degrading performances \cite{lunze2010sfa} and with stability and robustness guarantees \cite{Marchand2013}. Numerous Event-Based control strategies in the literature are focusing on stability and performance guarantees. Most event-based PID controllers are based on level-crossing triggering of some measuring error (see for instance \cite{arzen1999seb,Durand2009c}) or more generally rely on an event-function based on Lyapunov functions (see for instance \cite{velasco2009ols,Marchand2013}).

The two introduced Event-Based control algorithms are: (i) Event-Based Learning Rate control, which will be implemented to prevent sudden drop of the learning rate when the model is approaching the optimum; (ii) Event-Based Learning Epochs control, which will decide based on the learning speed when to switch to the next data batch.

Our algorithm is evaluated on two classical machine learning image datasets CIFAR-10 and CIFAR-100~\cite{Krizhevsky09}. The results are compared with four best state-of-the-art algorithms: Adam, Nadam, AMSGrad and AdaBound. Our results show that the E/PD combined with the two introduced Event-Based control not only outperforms original E/PD but also converges faster than any other state-of-the-art counterpart.

The article is organised as follows: after a brief introduction of the problem in Section~\ref{sec:Introduction}, we detail the scenario and the system to be controlled (i.e. a CNN) with its input and output metrics in Section~\ref{sec:Background}. The contribution, i.e., the two event-based mechanisms, is described in Section~\ref{sec:Control}.
Section~\ref{sec:Evaluation} contains the experimental setup, results and analysis. 
The article ends with a conclusion and perspectives for further work in Section~\ref{sec:Conclusion}.

\section{Background}
\label{sec:Background}



\vspace{-0.2em}
\subsection{Classical Online Learning Scenario}
\label{sec:classical_online}
\begin{figure}[htb]
	\centering
	\vspace{-0.3cm}
	\includegraphics[width=\linewidth]{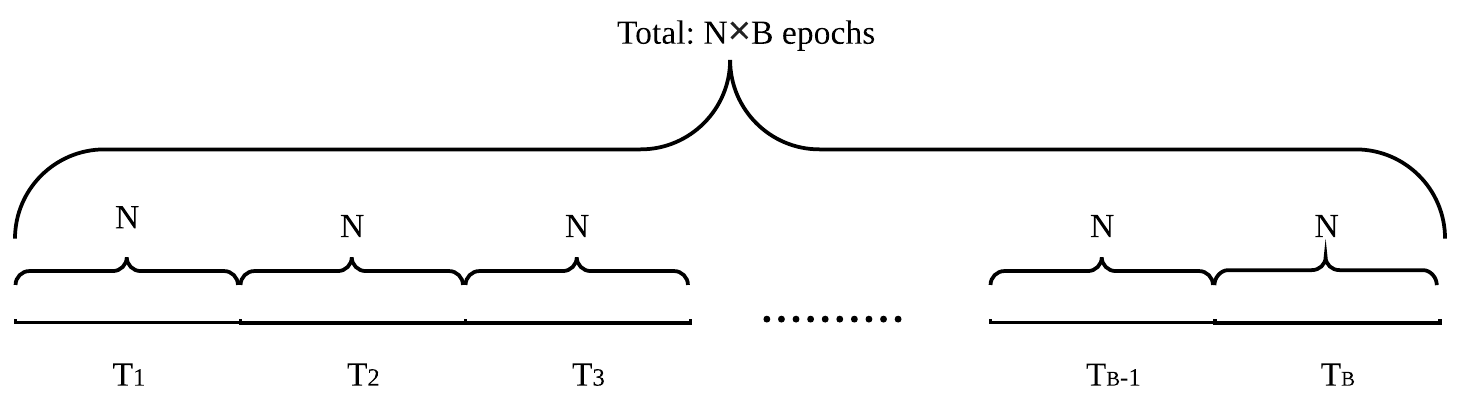}
	\vspace{-1.4em}
	\caption{Classical Online Learning Scenario.}
	\label{fig:classical_online}
	\vspace{-1.5em}
\end{figure}

We consider a dataset $\mathcal{T}$ with a total number of training instances $T$, each one belonging to a class $c: \mathbb{Z}^+ \to [1,C]$. The whole dataset is composed of $B$ subsets (i.e. batches), $T_i$ is the $i^{th}$ batch where $i: \mathbb{Z}^+ \to [1,B]$. Each batch equally contains $S$ data instances and will be used to train the model for $N$ epochs (i.e. $N$ times). At the reception of a new batch, the learning rate algorithm is reset with initial values. Classical online learning scenario is illustrated in Fig.~\ref{fig:classical_online}.

\subsection{Convolutional Neural Network and Gradient}
\vspace{-0.1em}
Convolutional Neural Network (CNN) is the state-of-the-art learning mechanism for image classification 
\cite{NIPS2012_krizhevsky}. 
CNN neurons functions are parameterized with weights and, eventually, bias.
The objective of the learning phase is to make iterative adjustments to these biases and weights to better fit the data. These weights in the CNN are usually updated using Stochastic Gradient Descent techniques (SGD):
\begin{align*}
    \small\theta_j = \small\theta_{j-1} - \lambda \frac{\small\partial L}{\small\partial \theta}
    \label{eq:gradient}
\end{align*}
where vector $\theta_j$ represents the weights vector computed at $j^{th}$ discrete time instant, 
$\lambda$ is positive and denotes the learning rate. $L$ is the loss function
. As we are always trying to minimize the loss function, we suppose that there exists an optimal solution of parameters $\theta_j^*$.  
\vspace{-0.2em}
\subsection{Performance Metrics}
There exists many metrics to evaluate the performance of a CNN model \cite{li2016performance}
, we used two of the most classical: classification accuracy 
and loss value. 

For evaluating, machine learning researchers typically prepare a testing dataset which will not be used during the training process. At the end of each training phase (called from now on epoch), the testing dataset is used to evaluate the model by measuring the classification accuracy and the loss value. Accuracy is defined as:
\begin{align}
    \small\mbox{Accuracy} = \frac{\small\mbox{Number of correct prediction}}{\small\mbox{Total number of prediction}}
\end{align}
The loss $L$ is defined as the difference between the predicted value by the model and the true value. The most common definition of $L$ used for classification problems is cross-entropy \cite{rubinstein1999}:
\begin{align}
\small L=-\frac{1}{\small V}\sum\limits_{p=1}^{\small V} \sum\limits_{q=1}^{\small C} \ y_{p,q} \log(\hat{y}_{p,q}) + (1-y_{p,q}) \log(1-\hat{y}_{p,q})
\vspace{-0.3em}
\end{align}
where $V$ is the size of testing dataset and $C$ is the total number of classes and also the length of the prediction vector which is a probability vector.
\;$\hat{y}_{p,q}$ denotes the $q^{th}$ bit value of prediction vector for data sample $p$ while $y_{p,q}$ is the ground truth, indicating if data $p$ belongs to class $q$ ($y_{p,q} = 1$) or not ($y_{p,q} = 0$).

\section{Event-Based Control Laws}
\label{sec:Control} In \cite{zhao:hal-02115916}, an E/PD control of the learning rate is proposed consisting of an increasing phase followed by a PD phase. However, if an increase of the performance can be achieved on both the loss and the accuracy, the learning rate is progressively decreased by the E/PD control in the PD phase, even though a larger value of learning rate would be more efficient in term of performance. Since event-based PID have shown to be more efficient in terms of convergence \cite{Durand2009c}, we propose here to implement an event-based E/PD controller to control the learning rate. \cite{zhao:hal-02115916} also shows that significant improvements in terms of accuracy and loss only occurred at the first epochs of training each data batch, so after this stage there is no limited interest into continuing the learning on further epochs. Therefore, we propose a second event-based control to adapt the data batch loading process. 

\subsection{Event-Based Learning Rate}
\label{sec:Control_Learning_Rate}

\begin{figure}[htb]
	\centering
	\vspace{-1em}
	\includegraphics[width=\linewidth]{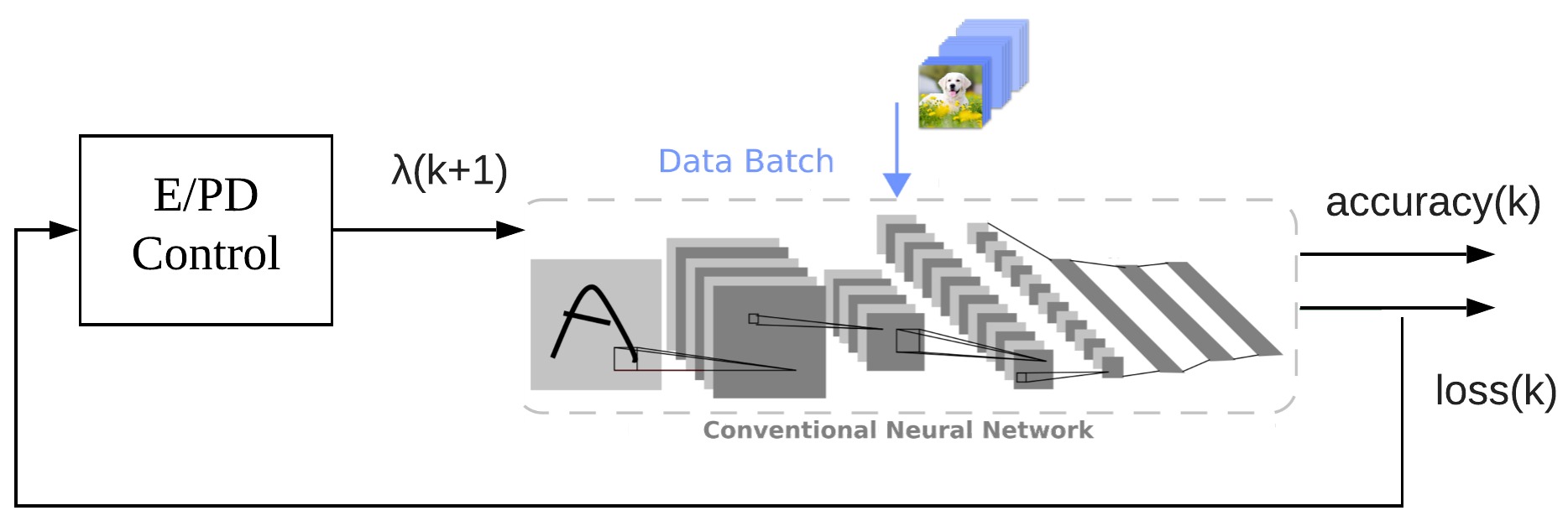}
	\caption{E/PD Control Structure.}
	\label{fig:eventbased_pd_framework}
	\vspace{-1em}
\end{figure}

A recall of the E/PD Control algorithm from \cite{zhao:hal-02115916} is schematicly presented in Fig.~\ref{fig:eventbased_pd_framework}. We suggest to look at a CNN training as a dynamical system with the learning rate as controlled input and the loss as measurable output. Initial weights of the CNN are chosen randomly and the initial learning rate $\lambda(0)$ is fixed. 
E/PD learning rate strategy is defined as:
\vspace{-0.2cm}
\begin{align}
    \lambda (k+1) = 2\lambda (k)  
\end{align}
as long as $L(k)<L(k-1)$ (E phase) and
\begin{align}
    \lambda (k+1) = 
    	K_P \dfrac{L(k)}{L(0)} - K_D \dfrac{L(k) - L(k-1)}{L(0)}  
\end{align}
from the first instant $k=k^*$ when $L(k^*)>L(k^*-1)$ to the end of learning process for the data batch (i.e. the PD phase). For the sake of simplicity the loss values are normalized with respect to the initial epoch loss value $L(0)$. $K_P$ and $K_D$ are the proportional and derivative gain detailed in \cite{zhao:hal-02115916}. 

On top of the PD phase we consider the following event base mechanism where instead of letting the PD-Control compute the rate each time (which might be lowering the learning rate), we propose to update the learning rate only if the loss value increases during the PD-Control phase. 

Let us define the event function $e_1: \R^+ \to \set{0,1}$ by:
\begin{equation} \label{event}
	e_{1_k} = \left\{
	\begin{array}{rl}
		1 & \mbox{if  } L(k) - L(k-1) >0\\
		0 & \mbox{otherwise} 	 
	\end{array} \right.
	\vspace{-0.3em}
\end{equation}
The proposed PD event-triggered control output $\lambda (k+1)$ at time $k+1$ is then: 
\begin{equation} \label{ebmpcU}
\scriptstyle
	\lambda (k+1) = \left\{
	\begin{array}{cl}
		\scriptstyle K_P \dfrac{\scriptstyle L(k)}{\scriptstyle L(0)} - \scriptstyle K_D \dfrac{\scriptstyle L(k) - L(k-1)}{\scriptstyle L(0)}  & \mbox{\small if  $e_{1_k}=1$} \\
		\scriptstyle \lambda (k) & \mbox{\small otherwise} 	 
	\end{array} \right.
\vspace{-0.3em}
\end{equation}
where $\lambda(k+1)$ is the calculated learning rate for epoch $k+1$, $L(k)$ is the corresponding loss for epoch $k$.

Note that the stability of CNN is ensured by E/PD, whose stability analysis is provided in \cite{zhao:hal-02115916}. Proposed event-based control does not introduce any instability because if $e_1 = 0$, which means the loss is decreasing, model is converging, and if $e_1 = 1$, the learning rate strategy returns to E/PD.



\vspace{-0.4em}
\subsection{Event-Based Learning Epochs}
\label{sec:Control_Learning_epoch}
\subsubsection{Controller Design}
\label{sec:eb_le_design}
As observed in \cite{zhao:hal-02115916}, significant improvement in the learning only occurs at the beginning when loading a new batch, the accuracy and loss value evolve slowly afterwards. This motivates the use of an event-based strategy on the loss value record.


Consider a maximum of $N$ training epochs within each batch. 
Let $X_k$ vector contains the latest $m$ epochs numbers
and $Y_k$ vector contains the $m$ latest corresponding normalized loss values:
\begin{align*}
 &X_k = \begin{bmatrix} k-m & \cdots & k-2 & k-1 & k \end{bmatrix}\\
 &Y_k = \begin{bmatrix}  \frac{L(k-m)}{L(0)} & \cdots & \frac{L(k-2)}{L(0)} & \frac{L(k-1)}{L(0)} & \frac{L(k)}{L(0)} \end{bmatrix}
\end{align*}
where $k \in [1, N-1]$. 
One can use least squares estimation to fit a regression line with $X_k$ and
$Y_k$:
\vspace{0cm}
\begin{align}
Y_k = \alpha_k X_k + \beta_k
\vspace{-0.3cm}
\end{align}
The purpose of this is that if the training process goes well the loss value should always decrease, therefore $\alpha_k$ should always be negative. Even with the presence of loss variations during the training, as long as the decreasing trend doesn't change, $\alpha_k$ should still be negative. Nevertheless, in the moment the loss trend becomes flat or even is increasing, $\alpha_k$ will become 0 or positive. 

We define the event mechanism by the event function $e_2: \R^+ \to \set{0,1}$ by:
\begin{equation} \label{event}
	e_{2_k} = \left\{
	\begin{array}{rl}
		\small \mbox{call new batch} & \mbox{\small if  } \scriptstyle\alpha_k \scriptstyle{>} \scriptstyle{\alpha_{thld}} \scriptstyle\mbox{ or } \scriptstyle{k=N} \\
		\small\mbox{remain on same batch} & \mbox{\small if } \scriptstyle\alpha_k \scriptstyle\leq \scriptstyle{\alpha_{thld}} \scriptstyle\mbox{ and } \scriptstyle{k<N} 	 
	\end{array} \right.
\end{equation}
which enables to switch to new data batch when the learning speed is too low, i.e. the training is not efficient anymore.

The threshold $\alpha_{thld}$ can be adjusted in order to control the efficiency of learning. This threshold should never be positive as an increasing curve of the loss value is not desirable. With enough computing resources and no time constraints, the threshold can be set close to $0$, and the training will last even though it makes very small improvement. Nevertheless, for online learning the time interval between two data batches can be short compared to the training time and we could encounter the scenario when before we finish the current training epochs the next data batch is already available. In this case, cutting off some useless training can be very useful. Therefore $\alpha_{thld}$ should also be chosen depending on the frequency of batch arrival. 
The choice of $m$ is based on the constraints imposed by the CNN (or the application using CNN). A large value of $m$ would imply a long time of inactivity as the controller would react only after $m$ epochs (consecutive tests). A small value of $m$ would imply that the algorithm is very sensitive to each epoch thus if $m=0$ the event based algorithm becomes a time based one. 

\subsubsection{Online Learning Scenario}
\label{sec:eb_le_online}
\begin{figure}[htb]
	\centering
	\vspace{-0.2cm}
	\includegraphics[width=\linewidth]{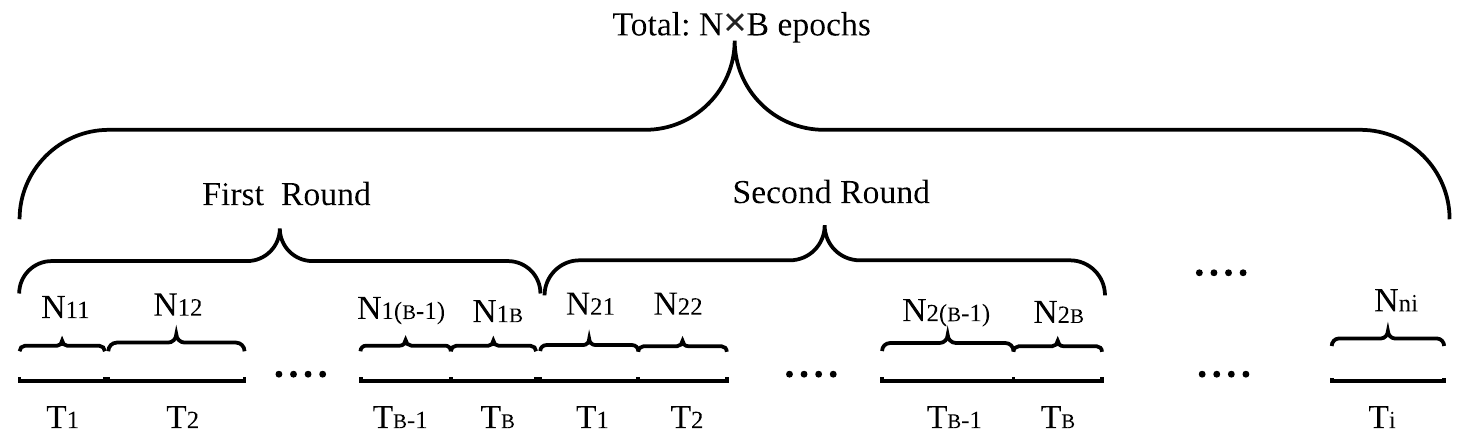}
	\caption{Event-Based Learning Epochs Online Learning Scenario.}
	\label{fig:eb_online}
	\vspace{-2em}
\end{figure}

Recall the online learning scenario defined in Sec.~\ref{sec:classical_online} and Fig.~\ref{fig:classical_online}, the difference for Event-Based Learning Epochs is that the training epochs for each batch could be varied but no larger than $N$, but the total training epochs are the same for both scenario for all the experiments of the same dataset. So here we could cyclically learn the data batches until it reaches the total epochs limit. The online learning arrangement for Event-Based Learning Epochs is illustrated in Fig.~\ref{fig:eb_online}.

\section{Experimental Evaluation}
\vspace{-0.3em}
\label{sec:Evaluation}
\subsection{Experimental Setup}
\vspace{-0.3em}

The experiments are implemented on two state of the art machine learning datasets: 1) CIFAR-10 (a natural image data set with 10 categories) and 2) CIFAR-100 (a natural image data set with 100 categories)~\cite{Krizhevsky09} with 3 different initial learning rate. 
The characteristics of the two data-sets are given in Table~\ref{tab:datasets}.
As the CIFAR-100 dataset has more classes, we use a deeper CNN: ResNet \cite{He2016DeepRL} than the one used for CIFAR-10 VGG \cite{simonyan2014very}. Due to the computational resource limitation, for ResNet with CIFAR-100, we train 30 epochs per data batch instead of 60 for CIFAR-10. 

\begin{table}[h]
\vspace{-0.2cm}
	\begin{center}
		\caption{Experiments configuration }
		\label{tab:datasets} \vspace{-0.1cm}
		\begin{tabular}{L{3.5cm} C{2cm} C{2cm}}
			\toprule
			\scriptsize\textbf{Use case}	& \scriptsize\textbf{CIFAR-10}	& \scriptsize\textbf{CIFAR-100} \\
			\midrule
			\scriptsize{\#data instances to train T}	& \scriptsize50,000					& \scriptsize50,000 \\
			\scriptsize\#data instances to test V	& \scriptsize10,000					&\scriptsize 10,000 \\
			\scriptsize \#classes $\textit{C}$ 		&\scriptsize 10 						&\scriptsize 100  \\
			\scriptsize data batch size $S$ 	&\scriptsize 10000					    &\scriptsize 10000   \\
		    \scriptsize total batches $B$ 	&\scriptsize 5	 & \scriptsize5   \\
			\scriptsize \#trainng epochs per batch $N$ 	&\scriptsize 60					    &\scriptsize 30   \\
			\bottomrule
		\end{tabular}
	\end{center}
	\vspace{-0.6cm}
\end{table}

All the experiments are implemented with Keras \cite{chollet2015keras} and are carried out on Google Cloud Compute-Engine using 8 virtual CPU with 30 GB memory and one P100 GPU. Each experiment is repeated 5 times.

The parameters $\alpha_{thld}$ and $m$ are selected through a process of cross validation on a subset of CIFAR-10. As a small value for $m$ leads to high sensitivity and a large $m$ slows down the detection of the situation, we predefined a reasonable list of choice $m \in [4;5;6;7;8]$. Due to similar consideration of sensibility, we also predefined a list for the learning rate threshold $\alpha_{thld} \in [-0.1; -0.01; -0.001; -0.0001]$. Each possible pair from these two lists is tested, a good compromise between reactivity and noise sensitivity was found for $m=4$ and $\alpha_{thld} = -0.001$.

\vspace{-1.0em}
\subsection{Evaluation Metrics}
\vspace{-0.2em}
The final loss and final validation accuracy (hereinafter  referred to as FVA) reveal the performance of the final model. Nevertheless, stability metrics are also important: if accuracy curve experiences a big variance near the end of training process, even we could have a good final result, we could not assure that we always get this result. Thus, in our evaluation, we include standard deviation of the accuracy of the last 10\% training epochs \cite{minaeemetrics} (hereinafter referred to as FASD (Final Accuracy Standard Deviation)). Convergence speed of accuracy is another metric to evaluate the performance, as we will focus on online learning scenario, the interval between two batch data can be short. With a limited time, a faster accuracy convergence could lead to a better model performance comparing to other algorithms. Therefore, we will report the first epoch when the experiment reaches the 95\% of best final accuracy among all the experiments.
\vspace{-1.0em}
\subsection{Evaluation of Event-Based E/PD}
\vspace{-0.2em}
Event-Based E/PD (hereinafter referred to as EB E/PD) refers to the E/PD control combined with Event-Based Learning Rate control (Sec.~\ref{sec:Control_Learning_Rate}).
We implement the online training experiments with E/PD and EB E/PD on CIFAR-10. 
From Fig.~\ref{fig:CIFAR10_double_eventbased_compare} we can first see the comparison between EB E/PD and original E/PD (only yellow and dotted blue line for now). For the first 60 epochs, we can see that EB E/PD is more stable than E/PD, then their curves are quite overlapped. The averaged comparison results are showed in Table.~\ref{tab:results_validation_cifar10}. EB E/PD performs better than E/PD in almost all metrics for all initial learning rate group. Even though EB E/PD has a higher FASD under 0.01 and 0.05 initial learning rate, but the minimum value of FVA($\pm$FASD) range of EB E/PD is higher than the maximum value of the range of E/PD.

\begin{table}[h!]
  \vspace{-0.2cm}
	\begin{center}
		\caption{Experiments with varying initial learning rate $\lambda(0)$ on CIFAR-10. Mean value over 5 runs are reported.}
		\label{tab:results_validation_cifar10}
		\begin{tabular}{L{1.3cm} C{0.7cm} C{0.7cm} C{2cm} C{1.8cm}}
			\toprule
			\scriptsize{\textbf{Algorithm}}	& \scriptsize{\textbf{$\lambda(0)$}}&\scriptsize{\textbf{Final loss}}	&\scriptsize{\textbf{$\mbox{FVA}^1$ ($\pm$ $\mbox{FASD}^2$) (\%)}}	 & \scriptsize{\textbf{1st epoch to $81.66\%^3$}} \\
			\midrule
			\scriptsize E/PD &\scriptsize0.002& \scriptsize0.58& \scriptsize{83.17($\pm$0.08)}	& \scriptsize{124/300} \\
			\scriptsize EB E/PD	& \scriptsize 0.002&\scriptsize \textbf{0.56 }& \scriptsize \textbf{83.81($\pm$0.03)} & \scriptsize \textbf{93/300}\\
			\midrule
			\scriptsize E/PD &\scriptsize 0.01& \scriptsize 0.55& \scriptsize 84.35($\pm$0.07)  &\scriptsize 88/300 \\
			\scriptsize EB E/PD	&\scriptsize0.01& \scriptsize\textbf{0.54 }&\scriptsize\textbf{84.91($\pm$0.10)}	& \scriptsize\textbf{75/300}\\
			\midrule
			\scriptsize E/PD &\scriptsize0.05& \scriptsize0.56&\scriptsize 85.06($\pm$0.12) &\scriptsize 73/300 \\
			\scriptsize EB E/PD	&\scriptsize 0.05& \scriptsize \textbf{0.50}&\scriptsize \textbf{85.96($\pm$0.26)} & \scriptsize \textbf{63/300}\\
			\bottomrule
			\multicolumn{5}{l}{\footnotesize 1. FVA: Final Validation Accuracy } \\
			\multicolumn{5}{l}{\footnotesize 2. FASD: Final Accuracy Standard Deviation}\\
			\multicolumn{5}{l}{\footnotesize 3. 81.66\%: 85.96\%(best final accuracy among all the experiments)$\times$95\%}
		\end{tabular}
	\end{center}
\end{table}

\begin{figure}[h]
	\begin{center}
	    \vspace{-1.55em}
		\subfloat[Loss value \vspace{-1em}]{
			\includegraphics[width=0.46\columnwidth]{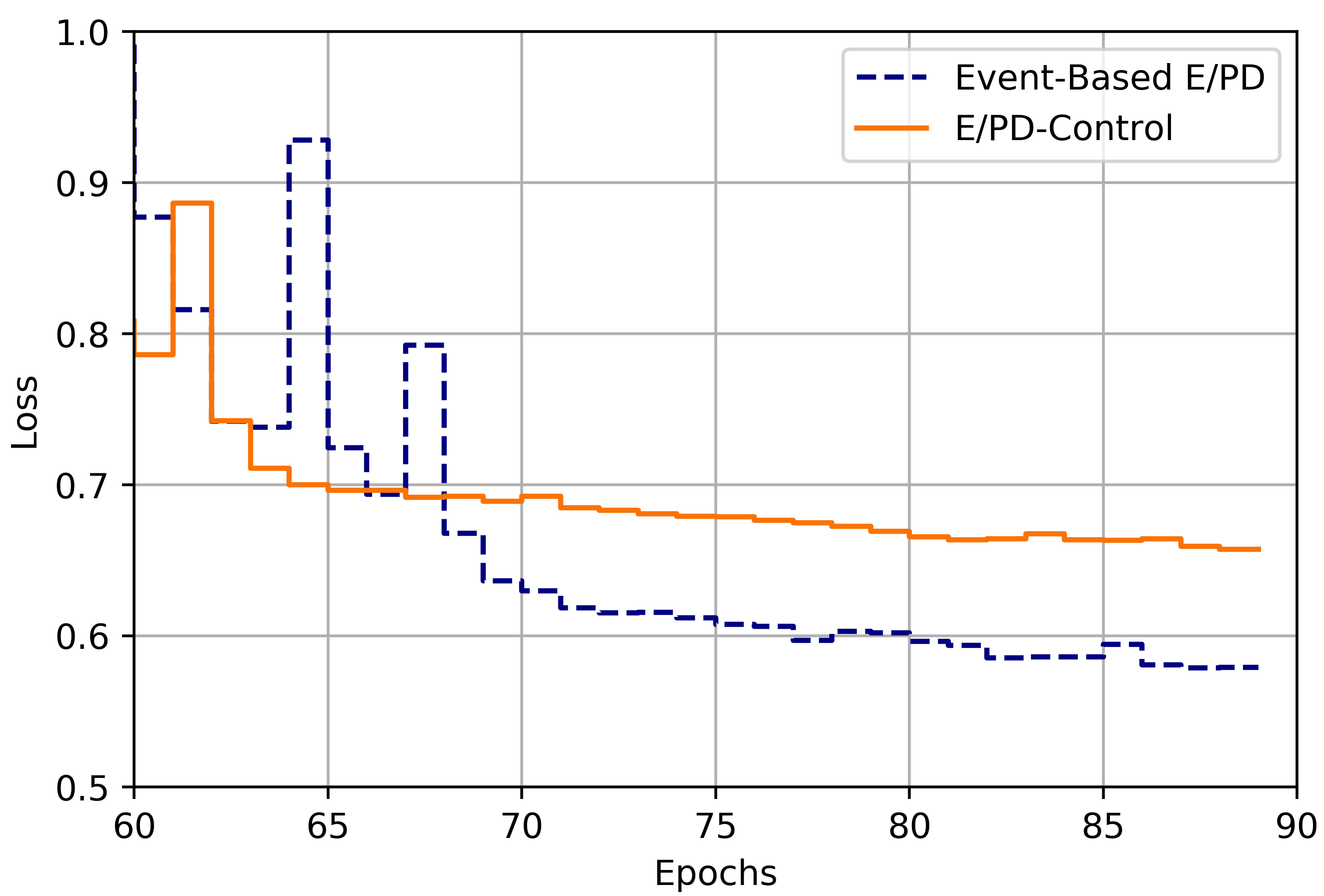}
			\label{fig:loss_compare_25}
		}
		\hfil
		\subfloat[Learning rate]{
			\includegraphics[width=0.48\columnwidth]{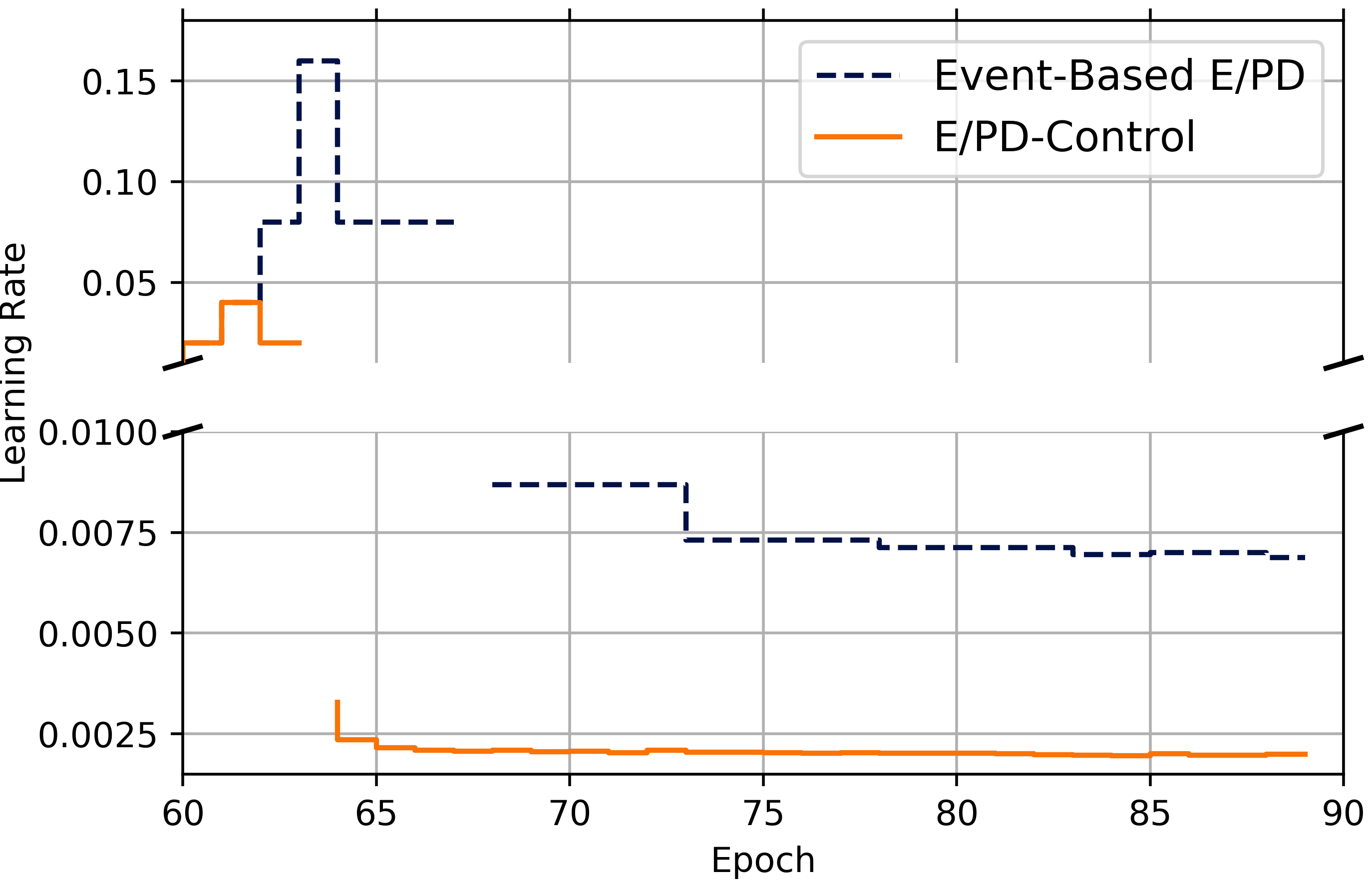}
			\label{fig:lr_compare_25}
		}
		\vspace{-0.3em}
		\caption{Performances of E/PD and EB E/PD on CIFAR-10}
		\vspace{-2.4em}
		\label{fig:CIFAR10_loss_lr_compare}
	\end{center}
\end{figure}

For the sake of visibility, we zoom into the 60th to 90th training epochs from our two experiment runs and show the evolution of the loss value and learning rate in Fig.~\ref{fig:CIFAR10_loss_lr_compare}. 
According to the learning rate curve, we know that E phase ends at 62th epoch for E/PD-Control curve, and at 64th epoch for EB E/PD.
E/PD-Control curve clearly shows the problem we mentioned above, we can observe that from 62th epoch, the loss of E/PD is continuously decreasing until 70th epoch, and its learning rate is also decreasing during this period. If the learning rate could stay constant during these 9 epochs, its loss would decrease sharply and that would improve the convergence speed. In contrast, EB E/PD keeps the learning rate when the loss continuously decreases which helps to accelerate the convergence. We can also notice that with the drop of the loss, each time when we update the learning rate for EB E/PD, its trend is also decreasing which will guarantee the stability of EB E/PD near the optimum.

\subsection{Evaluation of Double-Event-Based E/PD}\label{doubleEB_EPD}
\vspace{-0.15em}
Double-Event-Based E/PD-Control (hereinafter referred to as D-EB E/PD) refers to the E/PD control combined with Event-Based Learning Rate control (Sec.~\ref{sec:Control_Learning_Rate}) and Event-Based Learning Epochs control (Sec.~\ref{sec:Control_Learning_epoch}).
To ensure the need of the Event-Based Learning Rate control, we implemented E/PD with only Event-Based Learning Epochs control; results showed that Double Event-Based E/PD always has a better performance in Final loss and FVA. Due to the page limitation, we exclude these results from the main manuscript, however they are available online as appendices.

D-EB E/PD-Control has been tested on CIFAR-10 and CIFAR-100 and compared with 4 best state-of-the-art adaptive optimization algorithms: Adam, Nadam, AMSGrad and AdaBound. For these 4 learning rate strategies, except varying initial learning rate, all the other parameters remain as default as they mentioned in their paper or coded in Keras. As we adopt Event-Based Learning Epochs control into D-EB E/PD, the training epochs for each data batch is not fixed, we may also iterate each data batch several times. Therefore, we will not only report the results at the end of whole training process, but also the results after first round training (i.e. the training process iterates, for the first time, all the data batches, refer to Fig.~\ref{fig:eb_online}).

\begin{figure*}[h]
	\begin{center}
		\subfloat[Loss value \vspace{-1em}]{
			\includegraphics[width=1\columnwidth]{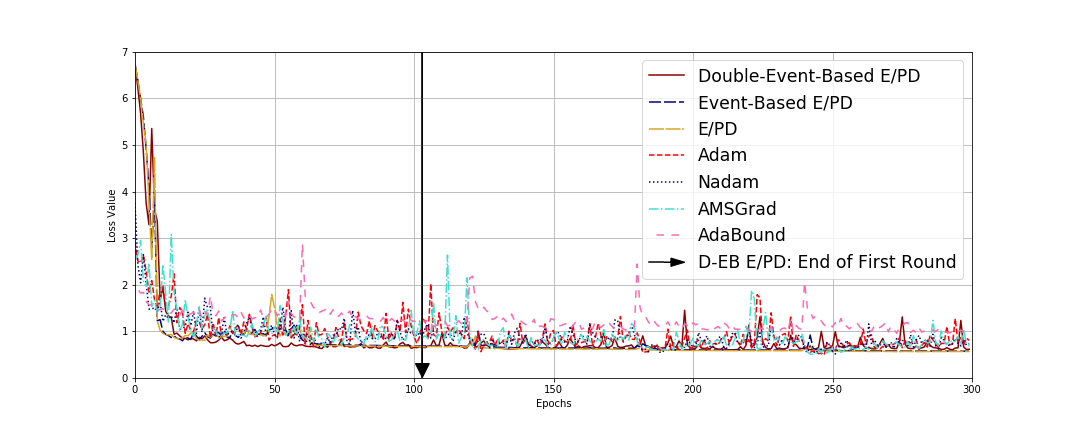}
			\label{fig:loss_compare_doubleevent}
		}
		\hfil
		\subfloat[Accuracy]{
			\includegraphics[width=1\columnwidth]{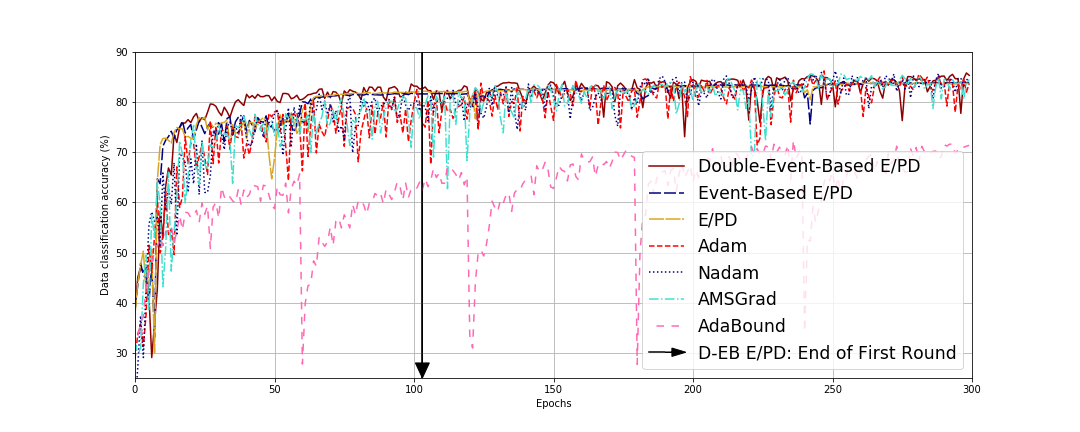}
			\label{fig:lr_compare_doubleevent}
		}
		\vspace{-0.3em}
		\caption{Performance comparison on CIFAR-10 with $\lambda(0)=0.01$ initial learning rate. Compact view of the results in Table~\ref{tab:results_CIFAR10_first_round}.}
		\vspace{-2.3em}
		\label{fig:CIFAR10_double_eventbased_compare}
	\end{center}
\end{figure*}

Experimental results on CIFAR-10 are showed in Fig.~\ref{fig:CIFAR10_double_eventbased_compare}, all the curves are generated with the same initial learning rate 0.01. Between 25th and 60th epoch, D-EB E/PD largely outperforms all the counterparts.
The vertical line with arrow at 104th epoch indicates that our D-EB E/PD algorithm has finished its first round learning of the whole 5 batches after this epoch. 
There are two reasons that we can achieve this performance: (i) EB E/PD converges very fast, (ii) during these epochs, our D-EB E/PD algorithm have trained with later batches data, while other 4 algorithms, they are still working on the first batch data. Diversity of training data helps to reach better performance. 

More detail of results on CIFAR-10 is reported in Table.~\ref{tab:results_CIFAR10_first_round}. 
D-EB E/PD reaches a higher final accuracy and lower final loss no matter $\lambda(0)$. Even though D-EB E/PD has a higher FASD than AdaBound with $\lambda(0)=0.01$ and $\lambda(0)=0.05$, the FVA($\pm$FASD) range of D-EB E/PD is always higher than the range of AdaBound. Additionally it only takes about 32 to 38 epochs to reach 95\% best accuracy in any group. All the indicators are very stable across different groups for D-EB E/PD. One can also note that for all the 4 state-of-the-art algorithms, they all perform very bad with $\lambda(0)=0.05$, they cannot even reach the 95\% best accuracy. We also implemented the same experiments with $\lambda(0)=0.25$. Except our algorithm, no other one reaches a reasonable accuracy value, which can be explained by the fact that during the PD phase of E/PD control our learning rate can decrease to a low level while the counterparts can not. Those results are available as appendices.

CIFAR-100 results are reported in Table.~\ref{tab:results_fashionmnist_first_round}. According to the FVA, we know that all the algorithms did not totally converge in the end of training process, but that does not influence our conclusion of analysis. D-EB E/PD outperforms other algorithms in almost all the metrics, when its FASD is higher than others in certain groups, its FVA($\pm$FASD) range is always higher than others. As the algorithms are not totally converged, the trend of accuracy curve is still increasing, therefore, the higher the initial learning rate, the faster the 1st epoch to reach 95\% best accuracy.

Table.~\ref{tab:firsround_result} shows the results of D-EB E/PD in the end of first round learning. All the final loss after first round learning in this table is lower than all the state-of-the-art algorithms in their end of whole training process comparing to their own group. Except CIFAR-100 for $\lambda(0)=0.002$, all the FVA after first round learning in this table exceed the 95\% best accuracy in Table.~\ref{tab:results_CIFAR10_first_round} and Table.~\ref{tab:results_fashionmnist_first_round}, respectively. As the learning process on CIFAR-100 is not totally converged, we can notice that the ending epoch of their first round is near the end of whole training process, our event-based control did not cut off many epochs. But for CIFAR-10, event-based control helps to massively cut off around 62\% to 67\% training epochs meanwhile guarantee a very good result.

\vspace{-0.6em}
\subsection{Trade-offs and limitations}
\vspace{-0.1cm}
The addition of event-based mechanisms improves the performance in terms of final accuracy and loss, however at the cost of two sacrifices: (i) Event-Based Learning Epochs accelerate the speed of learning each data batch. However, if we are not allowed to keep in cache any data batch locally, i.e. only allowed to learn each data batch once, the performance of Double Event-Based E/PD after first round is slightly worse than the performance after all the training epochs. (ii) Double Event-Based E/PD will cyclically learn all data batches, and it will need to load and unload data batch more times than classical online learning setting. Loading (unloading) data into (from) memory needs time. These are extra costs for Double Event-Based E/PD, however negligible compared to the computing intensity of CNNs.

Regarding the limitation of the presented D-EB E/PD, we identified one potential case for which our algorithm will fail: if the training data contains mislabeled data. These data will lead the model to converge to a wrong optimum, and as the algorithm minimizes faster the loss function, it will be faster over-fitting to the noisy data than other algorithms. However, this fail is caused by poor data selection, and is not specific to our algorithm. 

\begin{table}[t]
\vspace{0.4em}
	\begin{center}
		\caption{Double-Event-Based E/PD algorithm experiments with varying initial learning rate $\lambda(0)$ on CIFAR-10. Mean value over 5 runs are reported.}
		\label{tab:results_CIFAR10_first_round}
		\begin{tabular}{L{1.6cm} L{0.7cm} C{0.8cm} C{1.8cm}C{1.6cm}}
			\toprule
			\scriptsize{\textbf{Algorithm}}&\textbf{$\lambda(0)$}&\scriptsize{\textbf{Final loss}}& \scriptsize{\textbf{FVA $\pm$FASD}} 	&  \scriptsize{\textbf{1st epoch to $80.94\%^1$}}\\
			\midrule
			\scriptsize{D-EB E/PD}  &\scriptsize{0.002}& \scriptsize\textbf{0.58}& \scriptsize\textbf{84.50($\pm$0.59)}&\scriptsize\textbf{38/300}\\
			\scriptsize{Adam}	&\scriptsize0.002& \scriptsize0.73& \scriptsize84.14($\pm$1.34) &\scriptsize64/300\\
			\scriptsize{Nadam} 	&\scriptsize0.002& \scriptsize0.71&\scriptsize83.29($\pm$1.11)&\scriptsize66/300\\
			\scriptsize{AMSGrad}	&\scriptsize0.002& \scriptsize0.67&\scriptsize84.21($\pm$1.65)&\scriptsize65/300\\
			\scriptsize{AdaBound}	&\scriptsize0.002& \scriptsize 0.81 &\scriptsize 84.31($\pm$0.96) &\scriptsize 75/300\\
			
			\midrule
		    \scriptsize{D-EB E/PD} &\scriptsize0.01& \scriptsize\textbf{0.61}&\scriptsize\textbf{84.83($\pm$1.29)}&\scriptsize\textbf{37/300}\\
			\scriptsize{Adam}	&\scriptsize0.01& \scriptsize0.79&\scriptsize83.98($\pm$1.58) &\scriptsize64/300 \\
			\scriptsize{Nadam} 	&\scriptsize0.01& \scriptsize0.75&\scriptsize84.15($\pm$1.29)&\scriptsize65/300\\
			\scriptsize{AMSGrad}	&\scriptsize0.01& \scriptsize0.65&\scriptsize84.21($\pm$1.50)&\scriptsize72/300\\
			\scriptsize{AdaBound}	&\scriptsize0.01& \scriptsize 0.84 &\scriptsize 79.22($\pm$1.21) &\scriptsize -\\
			
			\midrule
			\scriptsize D-EB E/PD  &\scriptsize0.05& \scriptsize\textbf{0.60}& \scriptsize\textbf{85.20($\pm$3.14)}&\scriptsize\textbf{32/300}\\
			\scriptsize Adam	&\scriptsize0.05 &\scriptsize5.98&\scriptsize48.93($\pm$14.06)&\scriptsize-\\
			\scriptsize Nadam 	&\scriptsize0.05& \scriptsize7.74&\scriptsize42.27($\pm$13.95)&\scriptsize-\\
			\scriptsize AMSGrad	&\scriptsize0.05&\scriptsize2.69&\scriptsize59.74($\pm$12.43)&\scriptsize-\\
			\scriptsize{AdaBound}	&\scriptsize0.05& \scriptsize 1.03 &\scriptsize 71.49($\pm$1.65) &\scriptsize -\\
			
			\bottomrule
			\multicolumn{5}{l}{\footnotesize 1. 80.94\%: 85.20\%(best final accuracy among all the experiments)$\times$95\%}
		\end{tabular}
	\end{center}
\end{table}
\begin{table}[t]
\vspace{-1em}
	\begin{center}
		\caption{Double-Event-Based E/PD algorithm experiments with varying initial learning rate $\lambda(0)$ on CIFAR-100. Mean value over 5 runs are reported}
		\label{tab:results_fashionmnist_first_round}
		\begin{tabular}{L{1.6cm} L{0.7cm} C{0.8cm} C{1.8cm}C{1.6cm}}
			\toprule
			\scriptsize{\textbf{Algorithm}}&\textbf{$\lambda(0)$}&\scriptsize{\textbf{Final loss}}& \scriptsize{\textbf{FVA ($\pm$FASD) (\%)}} 	&  \scriptsize{\textbf{1st epoch to $46.56\%^1$}}\\
			\midrule
			\scriptsize{D-EB E/PD}  &\scriptsize{0.002}& \scriptsize\textbf{2.59}& \scriptsize\textbf{45.69($\pm$1.94)}&\scriptsize-\\
			\scriptsize{Adam}	&\scriptsize0.002& \scriptsize3.40& \scriptsize31.29($\pm$3.23) &\scriptsize-\\
			\scriptsize{Nadam} 	&\scriptsize0.002& \scriptsize3.18&\scriptsize35.66($\pm$3.35)&\scriptsize-\\
			\scriptsize{AMSGrad}	&\scriptsize0.002& \scriptsize3.13&\scriptsize35.38($\pm$4.02)&\scriptsize-\\
			\scriptsize{AdaBound}	&\scriptsize0.002& \scriptsize 3.29 &\scriptsize 39.87($\pm$4.42) &\scriptsize -\\
						\midrule

		    \scriptsize{D-EB E/PD} &\scriptsize0.01& \scriptsize\textbf{2.41}&\scriptsize\textbf{48.14($\pm$3.34)}&\scriptsize\textbf{111/150}\\
			\scriptsize{Adam}	&\scriptsize0.01& \scriptsize4.94&\scriptsize8.11($\pm$2.04) &\scriptsize- \\
			\scriptsize{Nadam} 	&\scriptsize0.01& \scriptsize4.55&\scriptsize9.70($\pm$2.32)&\scriptsize-\\
			\scriptsize{AMSGrad}	&\scriptsize0.01& \scriptsize4.79&\scriptsize8.16($\pm$0.50)&\scriptsize-\\
			\scriptsize{AdaBound}	&\scriptsize0.01& \scriptsize 3.51 &\scriptsize 30.98($\pm$3.08) &\scriptsize -\\
						\midrule
			\scriptsize D-EB E/PD  &\scriptsize0.05& \scriptsize\textbf{2.38}& \scriptsize\textbf{49.01($\pm$10.52)}&\scriptsize\textbf{100/150}\\
			\scriptsize Adam	&\scriptsize0.05 &\scriptsize4.72&\scriptsize2.64($\pm$0.58)&\scriptsize-\\
			\scriptsize Nadam 	&\scriptsize0.05& \scriptsize4.74&\scriptsize1.88($\pm$0.79)&\scriptsize-\\
			\scriptsize AMSGrad	&\scriptsize0.05&\scriptsize4.68&\scriptsize1.98($\pm$0.56)&\scriptsize-\\
			\scriptsize{AdaBound}	&\scriptsize0.05& \scriptsize 3.69 &\scriptsize 19.03($\pm$2.42) &\scriptsize -\\
			\bottomrule
			\multicolumn{5}{l}{\footnotesize 1. 46.56\%: 49.01\%(best final accuracy among all the experiments)$\times$95\%}
		\end{tabular}
	\end{center}
	\vspace{-1em}
\end{table}
\begin{table}[h!]
	\begin{center}
		\caption{Double Event-Based E/PD experiments on CIFAR-10 and CIFAR-100 in the End of First Round. Mean value over 5 runs are reported.}
		\label{tab:firsround_result}
		\begin{tabular}{L{1.3cm} C{0.8cm} C{1.3cm} C{1.2cm} C{1.7cm}}
			\toprule
			\scriptsize{\textbf{Dataset}}	& \scriptsize{\textbf{$\lambda(0)$}}& \scriptsize{\textbf{EE of $\mbox{FR}^1$}}	&\scriptsize{\textbf{FL after $\mbox{FR}^2$}}	 &\scriptsize{ \textbf{FVA after $\mbox{FR}^3$ (\%)}} \\
			\midrule
			\scriptsize CIFAR10 &\scriptsize0.002&\scriptsize 99/300&\scriptsize 0.60	&\scriptsize 82.47 \\
			\scriptsize CIFAR10 &\scriptsize0.01&\scriptsize 104/300&\scriptsize0.62 &\scriptsize82.36\\
			\scriptsize CIFAR10 &\scriptsize0.05&\scriptsize 113/300&\scriptsize 0.62 &\scriptsize 82.75 \\
			\midrule
        	\scriptsize CIFAR100 &\scriptsize0.002&\scriptsize 148/150&\scriptsize2.61	&\scriptsize 44.98\\
			\scriptsize CIFAR100 &\scriptsize0.01&\scriptsize 148/150&\scriptsize 2.44 &\scriptsize 48.04 \\
			\scriptsize CIFAR100 &\scriptsize0.05&\scriptsize 146/150&\scriptsize2.41 &\scriptsize 48.95\\
			\bottomrule
			\multicolumn{5}{l}{\footnotesize 1. EE of FR: End Epoch of First Round } \\
			\multicolumn{5}{l}{\footnotesize 2. FL after FR: Final loss after First Round}\\
			\multicolumn{5}{l}{\footnotesize 3. FVA after FR: Final Validation Accuracy after First Round}
		\end{tabular}
	\end{center}
	\vspace{-3.5em}
\end{table}
\vspace{-0.1em}
\section{Conclusion and future work}
\label{sec:Conclusion}
\vspace{-0.1em}
Due to the limitation of computing resource or short interval time between two data batches, convergence speed of the loss and accuracy becomes especially important for online learning. E/PD control is a powerful learning rate algorithm when training neural network on an online learning scenario. Based on E/PD, this paper proposes two algorithms: (i) Event-Based Learning Rate algorithm and (ii) Event-Based Learning Epochs algorithm. 

The new algorithm firstly introduces an Event-Based control on PD phase of E/PD, when the loss continuously decreases, we prevent the learning rate to decrease during this period. Second Event-Based control is implemented to inspect the record of the loss value. If the loss record has the tendency to increase, showing little learning efficiency, we will drop the rest learning epochs for current data batch.

Results show that Double-Event-Based E/PD can massively cut off training epochs, and even results in a lower loss value. For instance with CIFAR-10 dataset, it could save up to 67\% training epochs.

As the Event-Based Learning Epochs control is independent from learning rate algorithm and dataset, this work could be further extended by implementing this control with language, image and numeric datasets on time-based decay SGD, Adam, Nadam, AMSGrad and AdaBound learning rate algorithms, to prove that by simply adding this event-based control, all the learning rate algorithms on any dataset can improve their performance on online learning scenario.

\bibliographystyle{IEEEtran}
\vspace{-0.1cm}
\bibliography{root}
\end{document}